\documentclass[a4paper, 10pt, conference]{cssconf}
\IEEEoverridecommandlockouts    
\usepackage{epsfig} 
\usepackage{amsmath} 

\title{\LARGE \bf
Restoration of Reduced Self-Efficacy Caused by Chronic Pain through Manipulated Sensory Discrepancy 
}

\author{Matti Itkonen,  Riku Kawabata, Satsuki Yamauchi, Shotaro Okajima, Hitoshi Hirata, and Shingo Shimoda
\thanks{M.I. is with the School of Computing, University of Eastern Finland, (corresponding author to provide e-mail: {\tt\small matti.itkonen@uef.fi}).}
\thanks{R.K., S.Y., S.O., H.H., and S.S. are with Graduate School of Medicine, Nagoya University, Japan}
}

\begin{document}
\maketitle
\thispagestyle{empty}
\pagestyle{empty}

\begin{abstract}
Human physical function is governed by self-efficacy, the belief in one's motor capacity. In chronic pain patients, this capacity may remain reduced long after the damage  causing the pain has been cured. Chronic pain alters body schema, affecting how patients perceive the dimension and pose of their bodies. 
We exploit this deficit  using robotic manipulation technology and augmented sensory stimuli through virtual reality technology.  We propose a sensory stimuli manipulation method aimed at modifying  body schema to restore lost self-efficacy.
\end{abstract}

\section{Introduction}
Chronic pain affects up to 30\% of world population~\cite{c1}, imposing massive burden to society. Pain mechanisms  necessary to protect human body from damage cause by both exogenous and endogenous threats. Unlike transient pain, which disappears as soon as the sensory stimuli ceases, or acute pain, which disappears after damage to pain sensors has been repaired, chronic pain persists beyond the expected recovery period. Pharmaceuticals alone cannot cure this complex condition, which is influenced by biological, psychological, and social factors~\cite{c1}. 

The human sensorimotor system can prepare and execute functional motions effortlessly. However,  \emph{self-efficacy}, a person’s belief in their motor capability, is lower in patients suffering from chronic pain~\cite{cse}. As a result, these patients experience limitations in motor function despite the absence of physiological restrictions.

The internal representation of body pose and dimensions, known as \emph{body schema}, directly influences an individual's confidence in performing motor tasks. A distorted body schema is commonly associated with chronic pain. Patients tend to exaggerate the dimensions of affected body parts and rely less on kinesthetic sense on the affected side~\cite{cbodypresent}.

The difference between the perceived and objective reality has been found to be helpful in the rehabilitation of chronic pain.  For instance, patients with phantom limb pain experience sensations from a limb that has been amputated. Therapies using mirrors, rubber hands, or virtual reality can make the patient believe that the lost limb is still intact, triggering changes in the brain that facilitate rehabilitation~\cite{cvrphantom}.  Similarly, virtual therapy for post-traumatic stress disorder uses virtual reality to  engage patient with a combination of uncomfortable and pleasant stimuli~\cite{cvposttrauma}.

The same cerebral and cerebellar brain areas activated during physical action are also activated during action simulation, which in its active form is known as motor imagery and in its passive form as action observation. Due to this tight connection, both forms of action simulation are utilized in motor learning and rehabilitation. Chronic pain patients show abnormal responses to action simulation as observed through brain imaging~\cite{c5}. The planning, execution, and observation of action activate the same neural circuits, thereby playing a supplementary role in recovery.

Our aim is to create a discrepancy in proprioception, a difference between perceived and actual body positions. By establishing a method to modulate a patient’s body schema in relation to their actual body pose, we can manipulate self-efficacy to overcome disease-caused restrictions. This study establishes a theoretical framework to propose a system capable of altering self-efficacy in chronic pain patients.

\begin{figure}
\centering
\includegraphics[width=5cm]{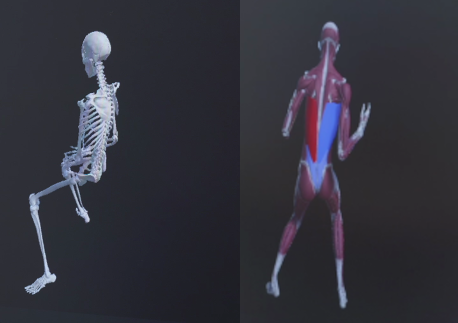}
 \caption{Skeletal and muscular avatars visualize real-time body responses for patients undergoing robotic rehabilitation. These visualizations aim not only to enhance the patient's awareness of anatomical and physiological dynamics that are otherwise invisible to them, but also to create subconscious feedback loops for improvement. The skeletal avatar is created using inertial measurement units (Xsens Technologies, Enschede, Netherlands), while the muscular avatar animates muscles using colors mapped to the magnitude of averaged surface electromyography data. This data is streamed via the Lab Streaming Layer (https://github.com/sccn/labstreaminglayer) and the visualizations are implemented using the Unity Game Engine (Unity Technologies, California, USA).}  
\label{segovia}
\end{figure}

\section{Method}
The human sensory system must collect afferent sensory signals from different modalities to make sense of the environment and itself in relation to the environment. These sensory signals are often redundant, noisy, and sometimes incongruous. To create a consistent perception, some afferent signals are amplified while others are attenuated. As a result of this process, some modalities may be ignored. Exploiting this mechanism can create a perception that is surprisingly different from objective reality (\emph{Sensory illusion}).  
  
Traditionally, an \emph{avatar} presents how others see a person in virtual reality. Fig. 1 illustrates an interactive avatar. A self-avatar, on the other hand, represents how a person sees themselves in virtual reality. Under favorable conditions, individuals can establish a strong emotional connection with their own avatar. Modulating the visible properties of the self-avatar can influence a person’s behavior and attitudes~\cite{cavatar} . 

The allure of virtual reality lies in its ability to create immersive, multi-sensory experiences. The concept of \emph{immersion} gauges the success of these experiences, heavily relying on the technological delivery of sensory stimuli. Inconsistencies in stimuli, such as uncontrolled haptic sensations conflicting with visual cues, can diminish immersion. In the worst cases, mismatches between modalities, like vision and vestibular senses, can lead to motion sickness, thereby limiting prolonged exposure.

In the context of avatars, \emph{embodiment} refers to a person's feeling that the avatar represents their own body. Achieving a higher degree of immersion is crucial for increasing embodiment.

In the \emph{rubber hand illusion}~\cite{cvrphantom}, a person is induced to believe that a rubber hand is part of their own body. The person sees a touch applied to the rubber hand, which is positioned close to their own hand. Simultaneously, the person's actual hand is touched. Once the embodiment is established, the person perceives touches to the rubber hand as if their actual hand were being touched. Smooth, slow touches on hairy skin, known as \emph{affective touch}~\cite{palpation}, have been found to strengthen embodiment.

\section{Proposal}
To establish an environment for modulating self-efficacy, we aim to induce the rubber hand illusion, enabling independent manipulation of kinesthetic and visual stimuli. We propose the use of augmented reality technology to blend sensory stimuli of both authentic and synthetic origins. Haptic stimuli, including tactile and kinesthetic feedback, will be generated using robotic devices. Visual sensory information will be provided through virtual reality technology.  

We present the patient with a self-avatar in an egocentric view using a high-end virtual reality headset to enhance immersion, based on the technology depicted in Fig. 1. The avatar must be highly responsive to kinematic displacements of the limb to promote embodiment. Subsequently, the limb is connected to a robotic device, which is also visualized to the patient. We anticipate that haptic feedback from the points where the limb interfaces with the robot during movement will contribute to a strong sense of embodiment. This effect may be enhanced by incorporating affective touch and simultaneous visual stimuli.

From this point onward, we initiate robot-assisted motion patterns targeting the joints of the affected limb. However, the visualization of the affected limb that the patient sees is altered in position, scaled to either a larger or smaller range of motion than reality, as illustrated in Fig. 2. The effectiveness of this method stems from factors such as immersion~\cite{cavatar}\cite{cvposttrauma}, embodiment~\cite{cvrphantom}, and action simulation~\cite{c5}. Additionally, patients receive feedback from physiological processes that they may use for pain reduction (Fig. 1).

\begin{figure}
\centering
\includegraphics[width=5cm]{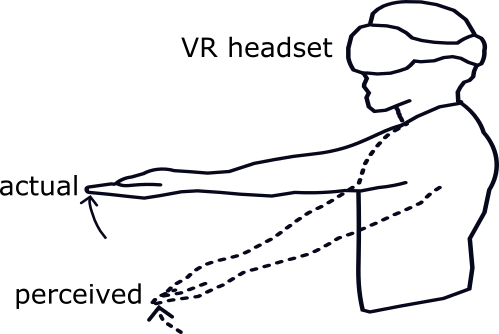}
 \caption{Modulation of limb perception in virtual reality involves the patient observing an embodied avatar of the affected arm. This avatar is animated based on the subject's actual motion, adjusted to an altered range of shoulder flexion.}  
\label{segovia}
\end{figure}

\section{Discussion}
The proposed method targets extending unconscious limits the patients body has set itself by giving misleading cues to multi-sensory integration. It uses action stimulation together with action to adjust body schema closer to reality. The parameterization, how the motion patterns (unimanual, bimanual, synchronous or asychnronous~\cite{biman}), mode of robotic assistance (active or passive) and the direction of altered visualization  (exaggeratement or understatement) will depend on patient segments. The presence of therapist will be necessary to adjust the various parameters  as well as dosing for each patient. 
 Initially, we are planning to target  patients with limited ability to control upper limbs due to musculoskeletal chronic pain to scope the suitability of the  method.

\end{document}